\documentclass[runningheads]{llncs}

\usepackage{microtype}
\usepackage{subcaption}

\usepackage{hyperref}
\usepackage{bbm}
\usepackage{multirow}
\usepackage{booktabs}       % professional-quality tables
\usepackage{amsfonts}       % blackboard math symbols
\usepackage{nicefrac}       % compact symbols for 1/2, etc.
\usepackage{microtype}      % microtypography
\usepackage{xcolor}         % colors
\usepackage{tikz}
\usetikzlibrary{arrows}
\usetikzlibrary{positioning,chains,arrows}
\usetikzlibrary{decorations.pathmorphing}
\usetikzlibrary{decorations.markings}
\usetikzlibrary{decorations.pathreplacing}
\usetikzlibrary{calc}
\usetikzlibrary{arrows.meta}
\usetikzlibrary{bending}
\usetikzlibrary{intersections}
\usetikzlibrary{fadings}
\usetikzlibrary{babel}
\usetikzlibrary{shapes}
\definecolor{out-col}{RGB}{230,230,230}
\definecolor{H1-1-col}{RGB}{255,217,47}
\definecolor{H1-2-col}{RGB}{153,0,194}
\definecolor{H1-3-col}{RGB}{55,126,184}
\definecolor{H2-1-col}{RGB}{255,127,0}
\definecolor{H2-2-col}{RGB}{77,175,74}
\definecolor{H2-3-col}{RGB}{152,78,163}

\tikzstyle{tips}=[%
btip/.tip = {Latex[length=6, width=6]},
-btip]
\tikzstyle{neuralnetwork}=[%
draw=black,
node distance = 3mm and 18mm,
start chain = going right,
neuron/.style = {rectangle, draw=black,
	minimum size=6pt, inner sep=4pt,
	on chain},
annot/.style = {text width=4em, align=center,},
tips]

\usepackage[many]{tcolorbox}
\newtcolorbox{mybox}[2]{%
    tikznode boxed title,
    enhanced,
    colframe={#2},
    arc=0mm,
    boxrule=3pt,
    interior style={white},
    attach boxed title to top left= {yshift=-\tcboxedtitleheight/2, xshift=20mm},
    fonttitle=\bfseries,
    colbacktitle=white,coltitle={#2},
    boxed title style={size=normal,colframe=white,boxrule=0pt},
    title=\fontsize{30pt}{24}\selectfont{#1}}

\usepackage{tikz}
\usetikzlibrary{shapes}
\usetikzlibrary{decorations.pathreplacing}
\usetikzlibrary{arrows.meta}
\usetikzlibrary{calc}
\usetikzlibrary{shapes}

\makeatletter
\newdimen\@myBoxHeight%
\newdimen\@myBoxDepth%
\newdimen\@myBoxWidth%
\newdimen\@myBoxSize%
\newcommand{\SquareBox}[2][]{%
    \settoheight{\@myBoxHeight}{#2}% Record height of box
    \settodepth{\@myBoxDepth}{#2}% Record depth of box
    \settowidth{\@myBoxWidth}{#2}% Record width of box
    \pgfmathsetlength{\@myBoxSize}{max(\@myBoxWidth,(\@myBoxHeight+\@myBoxDepth))}%
    \tikz \node [shape=rectangle, shape aspect=1,draw=red,inner sep=2\pgflinewidth, minimum size=\@myBoxSize,#1] {#2};%
}%
\makeatother

\tikzset{font={\fontsize{25pt}{24}\selectfont}}
\tikzset{operator/.style={rectangle, draw, inner sep=0pt, minimum size=2cm}}
\tikzset{mytip/.tip={>[length=12, width=16]}}

\definecolor{PaleBlue}{rgb}{0,.55,.9}
\definecolor{PaleGreen}{rgb}{0,.7,.25}
\definecolor{DarkGreen}{rgb}{0,.6,.25}
\definecolor{RedPink}{rgb}{.9,0,.2}
\definecolor{Pink}{rgb}{.85,.35,.7}
\definecolor{Purple}{rgb}{.6,0,.75}
\definecolor{Orange}{rgb}{.9,.3,.05}
\definecolor{GoldUL}{rgb}{1,.76,.32}

\colorlet{attentionColor}{Orange}
\colorlet{charEmbedColor}{RedPink}
\colorlet{predEmbedColor}{Pink}
\tikzstyle{embed}=[%
  draw,
  #1,
  anchor=north,
  minimum width=.8cm,
  minimum height=1.6cm,
  inner sep=0pt,
  text=#1!65!black,
  font=\fontsize{25pt}{24}\selectfont,
  ]

\usepackage{amsmath} % for \boxed{}
\usepackage{mathtools,xparse}
\usepackage{graphicx}
\usepackage{xcolor}
\usepackage{algorithm}
\usepackage{algorithmic}
\usepackage{tikz,pgf}
\usetikzlibrary{matrix,shapes,arrows,positioning}
\usepackage{enumitem}
\usepackage{color}
\usepackage{colortbl}
\definecolor{Gray}{gray}{0.9}

\def\equationautorefname~#1\null{%
  Equation~(#1)\null
}

\usepackage[toc,page,header]{appendix}
\usepackage{minitoc}
\renewcommand \thepart{}
\renewcommand \partname{}

\usepackage{listings}
\usepackage{multirow}
\usepackage{definitions}

\newcommand{\citep}{\cite}
\newcommand{\citet}{\cite}

\newtheorem{expprob}{Problem E.}

\newtheorem{intprob}{Problem I.}

\begin{document}

\title{On the Relationship Between Interpretability and Explainability in Machine Learning}

\titlerunning{On the Relationship Between Interpretability and Explainability in ML}
\author{\textbf{Benjamin Leblanc}}
\authorrunning{B. Leblanc}
\institute{
Université Laval, Québec, Canada\\
\email{benjamin.leblanc.2@ulaval.ca}}
\maketitle              % typeset the header of the contribution
\begin{abstract}
Interpretability and explainability have gained more and more attention in the field of machine learning as they are crucial when it comes to high-stakes decisions and troubleshooting. Since both provide information about predictors and their decision process, they are often seen as two independent means for one single end. This view has led to a dichotomous literature: explainability techniques designed for complex black-box models, or interpretable approaches ignoring the many explainability tools. In this position paper, we challenge the common idea that interpretability and explainability are substitutes for one another by listing their principal shortcomings and discussing how both of them mitigate the drawbacks of the other. In doing so, we call for a new perspective on interpretability and explainability, and works targeting both topics simultaneously, leveraging each of their respective assets.

\end{abstract}

\section{Introduction}

Machine learning (ML) is used in a variety of fields with numerous applications, such as image recognition \cite{NIPS2012_c399862d}, sentiment analysis \cite{NIPS2014_1415db70}, and language translation \cite{2018arXiv180701784D}. In some areas such as the medical field, ML-assisted predictions or decisions can drastically impact human life. For example, breast cancer \cite{NAJI2021487} can be devastating if not diagnosed in time (or at all). Still, many events made it clear that not understanding the inner workings and decision process of a predictor could lead to unfortunate events: job and loan applications biased toward men \cite{blog4}, mortgage-approval biased toward white applicants \cite{blog6}, higher credit card limits for men \cite{blog5}, etc. As pointed out by Goodman and Flaxman \citet{Goodman_Flaxman_2017}: “If we do not know how ML [predictors] work, we cannot check or regulate them to ensure that they do not encode discrimination against minorities [...], we will not be able to learn from instances in which it is mistaken.”

Two schools of thought seem to have emerged from this need for knowledge: either focusing on explaining black-boxes \cite{DBLP:journals/jair/RasXGD22,DBLP:journals/csur/GuidottiMRTGP19}, or favoring interpretable predictors while completely discarding explainability \cite{DBLP:journals/natmi/Rudin19}. %Interpretability and explainability, in this sense, are often seen as substitutes for one another. 
Both approaches contain many flaws: interpretability can't provide all that there is to know about a predictor \cite{a23325bf-6a64-3bba-a051-ebe1b2dce874}, all the while being subjective \cite{DBLP:journals/corr/abs-2103-11251}, and interpretable predictors are, most of the time, harder to train than predictors in general \cite{10.1214/aos/1176347963}. On the other hand, there is an obligation to trust the explanations  \citet{DBLP:conf/fat/BordtFRL22} which, by definition, are necessarily wrong \cite{DBLP:conf/icml/MolnarKHFDSCGB20,DBLP:journals/natmi/Rudin19} and can provide more question than it answers \cite{DBLP:conf/naacl/JainW19}. As it is often understood that explainability reduces the need for interpretability, and \textit{vice versa}, the literature on their practical application and the mitigation of their drawbacks is disjoint.

In this work, we challenge the common belief that interpretability and explainability are substitutes for one another and argue that they actually are complementary, especially in that they mitigate the shortcomings of the other. Our first contribution is met by listing and discussing the principal drawbacks of explainability and interpretability raised in the literature. For our second contribution, we discuss how these limitations either disappear or diminish when considering both explained and interpretable predictors. Throughout the paper, we also show how both concepts differ in their very nature, thus necessarily aiming at different knowledge.

Many works critically discuss interpretability and its relationship with predictive performances \citet{DBLP:journals/natmi/Rudin19} or with ML family of predictors \citet{DBLP:journals/cacm/Lipton18}, and its conceptual differences with explainability \cite{DBLP:journals/cacm/Lipton18,DBLP:journals/corr/abs-2103-11251,DBLP:journals/corr/abs-2010-12016,Krishnan2020-KRIAIA-3,doshivelez2017towards}. Yet, it is the first time, up to our knowledge, that a reconciliation between interpretability and explainability via their complementarity is attempted and pleaded. In doing so, we call for a new perspective on interpretability and explainability and for works targeting both topics simultaneously, leveraging each of their respective assets in practical applications.

\section{Defining and Discussing the Concepts at Hand}

Cases occur where a lot of definitions are attempted for defining a single concept (such as for interpretability \cite{DBLP:journals/csur/GuidottiMRTGP19,DBLP:conf/dsaa/GilpinBYBSK18,9380482,Zhang2020ASO,DBLP:journals/air/GrazianiDCAYVNABPPLRDAM23,DBLP:journals/corr/abs-1806-00069,DBLP:journals/sigkdd/ZytekALBV22,DoshiVelez2017TowardsAR,DBLP:journals/isci/GactoAH11,DBLP:journals/kbs/SaeedO23,DBLP:journals/computer/AkataBRDDEFGHHH20,DBLP:journals/jbi/MarkusKR21}), many resembling each other. Thus, clearly defining the terms that are used or which paradigm is referred to is necessary for ensuring efficient transfers of ideas and avoiding misunderstandings.

A \textit{task} is an ML problem, whereas a \textit{domain} is a set of tasks sharing similar explanatory variables and response variables. The prediction (decision) process of a predictor corresponds to how it transforms a given input into an output.

The \textit{inputs} of a problem corresponds to the explanatory variables. More broadly, a \textit{feature} could be an input, or could emerge from the interaction between several inputs, but is characterized by being meaningful. Sex, gender, postal code, annual income, and number of cars are features, whereas pixels and sound files are not.

We distinguish ML \textit{models}, \textit{algorithms}, and \textit{predictors}. A predictor simply is a mathematical function. An algorithm tunes the \textit{parameters} and the form of a predictor for it to become better, given criteria, at tackling a given task. Algorithms can also build a predictor from scratch \cite{DBLP:journals/corr/abs-2209-03450,DBLP:books/wa/BreimanFOS84,DBLP:conf/icml/MarchandS01}. In the machine learning literature, the term model is often used loosely to refer interchangeably to an algorithm, a predictor, or a family of predictors. In the following, we retain the last definition: a machine learning model is a family of predictors. For example, a linear model is the space of predictors outputting a linear function of its inputs.

\subsection{Interpretability}\label{subsec:int}

We do not aim at attempting a new definition of interpretability in machine learning: we rather refer to what preexisting paradigm we stand in. Inspired by Rudin \cite{DBLP:journals/corr/abs-2103-11251}, we consider that a degree of \textit{interpretability} of a predictor on a given task is proportional to the capacity of a given person to understand its decision process simply and only by considering the predictor in itself.

Just as predictive performances, many metrics could be used for measuring such a degree. Here, a predictor whose degree of interpretability is high will be simply called \textit{interpretable}, whereas when it is really low, a \textit{black-box}. A person evaluating the interpretability of a predictor will be called a \textit{judge}. Since the decision process of a predictor (both the inputs and the output) encapsulates the notion of \textit{domain}, interpretability is domain-specific and therefore eludes definitions that would be too restrictive. Understanding the decision process implies understanding the role of each parameter and feature and how they interact with each other to generate a prediction. We call this aspect the \textit{transparency} of a predictor. Transparency is more easily obtained when the predictor is both \textit{sparse} and \textit{parsimonious}; that is, when the predictor has only a few non-zero parameters and when only a few features are retained by the predictor, respectively.

But interpretability is even more complicated than that. As pointed out by \citet{DBLP:journals/corr/abs-2103-11251}, characteristics such as monotonicity for a given variable, decomposability into sub-predictors, ability to perform case-based reasoning, usage of complex but well-known (e.g., Newton's laws of physics) relations, preferences among the choice of variables, etc. impact just how interpretable a predictor is. Note that this notion is human-based and therefore is subjective, making it even harder to objectively quantify or define interpretability.

\subsection{Explainability}

All in all, interpretability greatly differs from \textit{explaining} a predictor, where inherently hidden information concerning its decision process is presented to a given person. Explainability thus refers to the methods for creating these explanations. As stated by \citet{DBLP:journals/inffus/ArrietaRSBTBGGM20}: “[E]xplainability is associated with the notion of explanation as an interface between humans and a [predictor] that is, at the same time, both an accurate proxy of the [predictor] and comprehensible to humans.” 

Those explanations are simplifications of the reality and, in machine learning, often presented in the form of simplifications of the original predictor \cite{DBLP:journals/corr/abs-2006-11371,DBLP:conf/kdd/Ribeiro0G16,DBLP:conf/aaai/SubramanianPJBH18,MONTAVON2017211,DBLP:journals/corr/LakkarajuKCL17,Zhou2016InterpretingMV,DBLP:conf/nips/LundbergL17}. Explainability techniques can be separated into several different (sometimes overlapping) subgroups, involving a varied nomenclature: \textit{global} \cite{Apley2016VisualizingTE,DBLP:journals/corr/abs-1805-04755,doi:10.1080/07350015.2019.1624293,DBLP:conf/kdd/Hooker04a,DBLP:journals/jmlr/FisherRD19}, describing the average behaviour of a predictor; \textit{local} \cite{Goldstein2013PeekingIT,DBLP:journals/corr/abs-1805-08819,DBLP:journals/corr/abs-1711-00399,DBLP:conf/ppsn/DandlMBB20}, describing a predictor behaviour for a given example; \textit{model-agnostic} \cite{DBLP:conf/ismir/MishraSD17,DBLP:conf/iccv/FongV17,DBLP:conf/nips/DabkowskiG17,DBLP:journals/rjour/StaniakB18}, applied to any model; \textit{model-specific} \cite{DBLP:journals/nn/FeraudC02,DBLP:conf/iccv/0018CTYZLH0J21,DBLP:conf/iclr/KindermansSAMEK18,DBLP:conf/his/ChavesVT05,DBLP:journals/eor/MartensBGV07}, concerning a specific model; etc. This \textit{explanatory} use of explainability is different from its \textit{exploratory} purpose \cite{DBLP:conf/iclr/AtreyCJ20} (i.e. being used practically during the training of ML predictors); for example, feature selection \cite{DBLP:journals/electronicmarkets/ZachariasZCH22}. When conveying an explanation, the \textit{explainer} is the person transmitting it and the \textit{explainee} is the person receiving it.

\renewcommand{\arraystretch}{1.25}
\begin{table}
    \centering
    \begin{tabular}{|l|l|l|}
        \hline
        \multirow{2}{7em}{When is the problem faced?} & \multirow{2}{4.8em}{Number} & \multirow{2}{24em}{The problem itself}\\
        &  & \\
        \hline
        \hline
        \multirow{10}{7em}{Explainability} & Problem 1 & \textit{There is an obligation to blindly trust the explanation.} \\
        \cline{2-3} 
        & Problem 2 & \textit{Explanations are necessarily wrong.} \\
        \cline{2-3} 
        & \multirow{3}{4.8em}{Problem 3} & \multirow{3}{24em}{\textit{There necessarily is a misalignment between what the explainer wants to convey and what the explainee actually understands.}} \\
        &  & \\
        &  & \\
        \cline{2-3}
        & Problem 4 & \textit{Explanations can create more questions than it answers.} \\
        \cline{2-3}
        & \multirow{2}{4.8em}{Problem~5*} & \multirow{2}{24em}{\textit{Computing the explanations might involve huge computation time.}}\\
        &  & \\
        \cline{2-3}
        & Problem~6* & \textit{Most techniques are vulnerable to adversarial attacks.}\\
        \hline
        \multirow{6}{7em}{Interpretability} & \multirow{2}{4.8em}{Problem 1} & \multirow{2}{24em}{\textit{Interpretability can’t provide all that there is to know about a predictor.}}\\
        &  & \\
        \cline{2-3} 
        & Problem 2 & \textit{Interpretability is subjective.} \\
        \cline{2-3} 
        & Problem 3 & \textit{Enforcing interpretability can make training harder.} \\
        \cline{2-3} 
        & \multirow{2}{4.8em}{Problem 4} & \multirow{2}{24em}{\textit{Interpretability as a means is not enough to ensure some of its most common ends.}}\\
        &  & \\
        \hline
    \end{tabular}
    \caption{The principal limitations of explainability and interpretability that are discussed (raised*) in the article.}
    \label{tab:list_prob}
\end{table}

\section{The relationship between interpretability and explainability}\label{sec:intexp}

In this section, we make the case that interpretability and explainability actually are \textit{complementary} to one another. In order to do so, we will look at the flaws and limitations in the information that can be gathered both in explaining black-boxes (Subsection \ref{sec:flaws_exp}) and in interpretable predictors that are not explained in any way (Subsection \ref{sec:flaws_int}). The various points that are discussed are listed in \autoref{tab:list_prob}. Then, we discuss how these problems can be attenuated when it comes to explained interpretable predictors (Subsection \ref{sec:reconc_int_exp}). Note that even though the flaws we will be discussing are raised separately, they must be understood as a whole, for they are not independent. 

\subsection{The Flaws in Explaining Black-Boxes}\label{sec:flaws_exp}

\begin{expprob}\label{expprob_1}
    There is an obligation to blindly trust the explanation.
\end{expprob}

Explanations form a proxy between the user and something that is, by definition, inscrutable to human beings; they cannot be authentically verified but by other proxies. The explanations must therefore be trusted blindly, or substantiated with other explanations, where the same problem arises.

Some explainability techniques display interesting properties ensuring they behave as expected, but most of the time, and because explainability tends to be used on truly complex predictors, simplifications of the techniques (heuristics) are required, leading to cases where the properties do not even hold. This is notably the case for SHAP values \cite{DBLP:conf/nips/LundbergL17} computed with the TreeSHAP algorithm \cite{DBLP:journals/corr/abs-1802-03888}, which violates the \textit{missingness} property of SHAP values: "TreeSHAP was introduced as a fast, model-specific alternative to KernelSHAP, but it turned out that [...] features that have no influence on the prediction function \textit{f} can get a TreeSHAP estimate different from zero" \cite{molnar2022}.

In addition, there is an obligation not only to trust the explanation but also the explainer. The explanations might truly reflect what the explainer wanted to convey, but this has to be aligned with what information the explainee wants to get. As stated in \citet{DBLP:conf/fat/BordtFRL22}: “most situations where explanations are requested are adversarial, meaning that the explanation provider and receiver have opposing interests and incentives, so that the provider might manipulate the explanation for her own ends.”

\begin{expprob}\label{expprob_3}
    There necessarily is a misalignment between what the explainer wants to convey and what the explainee actually understands.
\end{expprob}

Miller \citet{DBLP:journals/ai/Miller19} gives precious insights on that matter: “[I]t is fair to say that most work in explainable artificial intelligence uses only the researchers’ intuition of what constitutes a ‘good’ explanation. There exist vast and valuable bodies of research in philosophy, psychology, and cognitive science on how people define, generate, select, evaluate, and present explanations, which argues that people employ certain cognitive biases and social expectations towards the explanation process\footnote{See Malle \citet{article_malle} when it comes to explanation selection, and Kahneman et al. \citet{Kahneman1982-KAHJUU}, Miller and Gunasegaram \cite{Miller_Gunasegaram_1990} and Girotto et al. \citet{GIROTTO1991111} when it comes to counterfactual explanations.}." Also: “[The explanations] are a transfer of knowledge, presented as part of a conversation or interaction, and are thus presented relative to the explainer’s beliefs about the explainee’s beliefs. This goes without discussing the misalignment between what the technique conveys and what the explainer actually wants to convey.” This is notably the case for the TreeSHAP algorithm \cite{DBLP:journals/corr/abs-1802-03888}, which defines its value function differently than in the original work on SHAP values \cite{DBLP:conf/nips/LundbergL17}, thus changing the interpretation of the obtained values, but which could easily be misunderstood or ignored by the explainer.

\begin{figure}[t]
    \centering
    \includegraphics[width=0.75\textwidth]{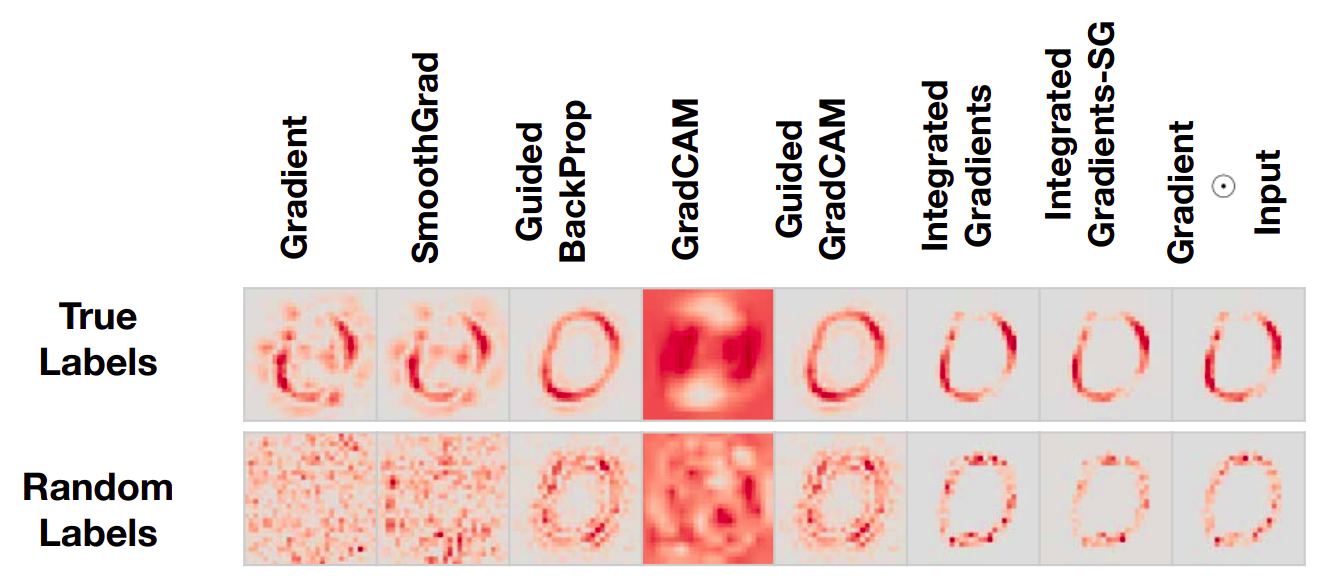}
    \caption{Borrowed from \citet{DBLP:conf/nips/AdebayoGMGHK18}, adapted: two predictors were trained on the MNIST task: one with the regular labels and one with random labels. Both predictors provide similar saliency maps for similar inputs even though we know for sure the second one hasn't learned anything.}
    \label{fig:random}
\end{figure}

Problems also arise on the explainee's part only: the various explainability techniques aiming at computing similar information (e.g. Permutation feature importance \cite{DBLP:journals/jmlr/FisherRD19} and Leave one covariate out \cite{doi:10.1080/01621459.2017.1307116}, both aiming at computing the same information) have been observed to yield important discrepancies, even on small tasks tackled by a simple sparse linear predictor \cite{markus2023understanding}. If the subtleties distinguishing these methods are out of reach for most explainees, then how do explainees account for the discrepancies they observe?

\begin{expprob}\label{expprob_2}
    Explanations are necessarily wrong.
\end{expprob}

The fact that explanations “condense the complexity of machine learning [predictors] into human-intelligible descriptions that only provide insight into specific aspects of the [predictor] and data” \cite{DBLP:conf/icml/MolnarKHFDSCGB20}, or that “if the explanation was completely faithful to what the original [predictor] computes, the explanation would equal the original [predictor], and one would not need the original [predictor] in the first place, only the explanation” \cite{DBLP:journals/natmi/Rudin19}. 

A black-box is likely to be a truly complex function (especially in the era of deep learning \cite{DBLP:conf/cvpr/SzegedyLJSRAEVR15}) and while many explanation techniques rely on extrapolating between training examples \cite{Shapley,DBLP:conf/nips/LundbergL17,DBLP:conf/pkdd/CasalicchioMB18,DBLP:journals/jmlr/FisherRD19,10.1214/aos/1176347963,10.1214/aos/1013203451,DBLP:journals/corr/abs-1805-04755}, this can easily lead to unreliable results \cite{DBLP:journals/corr/abs-2006-04628,DBLP:journals/sac/HookerMZ21,DBLP:conf/icml/MolnarKHFDSCGB20}. And even when explainability is used directly with training points, in many cases, assumptions are made on the data distribution (for example: assuming feature independence \cite{DBLP:conf/nips/LundbergL17,DBLP:conf/kdd/Ribeiro0G16,DBLP:journals/kais/StrumbeljK14,DBLP:conf/icml/ShrikumarGK17,DBLP:conf/sp/DattaSZ16}) or the predictor itself (for example: assuming linearity of the predictor \cite{DBLP:conf/nips/LundbergL17}). Such assumptions being most of the time unlikely to be met, this leads to unexpected results \cite{molnar2022} or in worse cases misleading or false characterizations \cite{DBLP:conf/ijcai/LaugelLMRD19,DBLP:journals/natmi/Rudin19,DBLP:conf/aies/LakkarajuB20}. 

Human intuition can be fooled and therefore is unreliable for deciding to trust an explanation, which is sometimes right but for wrong, surprising, or incoherent reasons \cite{DBLP:conf/nips/AdebayoGMGHK18}. %, which isn't more desirable than a transparent false explanation.
\autoref{fig:random} shows a speaking example of this behavior.

\begin{expprob}\label{expprob_4}
    Explanations can create more questions than it answers.
\end{expprob}

Rudin et al. \citet{DBLP:journals/corr/abs-2103-11251}, among others \cite{DBLP:conf/naacl/JainW19}, give an insightful example on that matter: “Saliency maps highlight the pixels of an image that are used for a prediction, but they do not explain how the pixels are used. As an analogy, consider a real estate agent who is pricing a house. A \textit{black-box} real estate agent would provide the price with no explanation. A \textit{saliency} real estate agent would say that the price is determined from the roof and backyard, but doesn’t explain how the roof and backyard were used to determine the price.” 

\autoref{fig:random} shows an example where anyone could get fooled by assuming both predictors to have learned how to recognize the shape of a "0" digit, but only knowledge of the training scheme reveals that it is not the case.
Here are two other problems (among others) that won't be discussed in detail, but that are worth raising:

\begin{itemize}
    \item \textit{Computing the explanations might involve huge computation time}; As raised in \autoref{expprob_1}, being faithful to the proposed in the literature might be a computational burden \cite{DBLP:journals/jmlr/FisherRD19,DBLP:journals/ress/WeiLS15a,DBLP:conf/nips/LundbergL17}, leading to approximations and heuristics which in turn can lead to \autoref{expprob_3};
    \item \textit{Explainability techniques are vulnerable to adversarial attacks}; not only when it comes to the values that are computed \cite{DBLP:conf/aies/LakkarajuB20,DBLP:conf/ijcai/LaugelLMRD19,DBLP:conf/aies/SlackHJSL20,DBLP:conf/aaai/GhorbaniAZ19,DBLP:journals/corr/abs-2205-15419}, but also in the sense instilled in the obtained values \cite{DBLP:conf/fat/BordtFRL22,13818}.
\end{itemize}

\begin{figure}[t]
    \includegraphics[width=\textwidth]{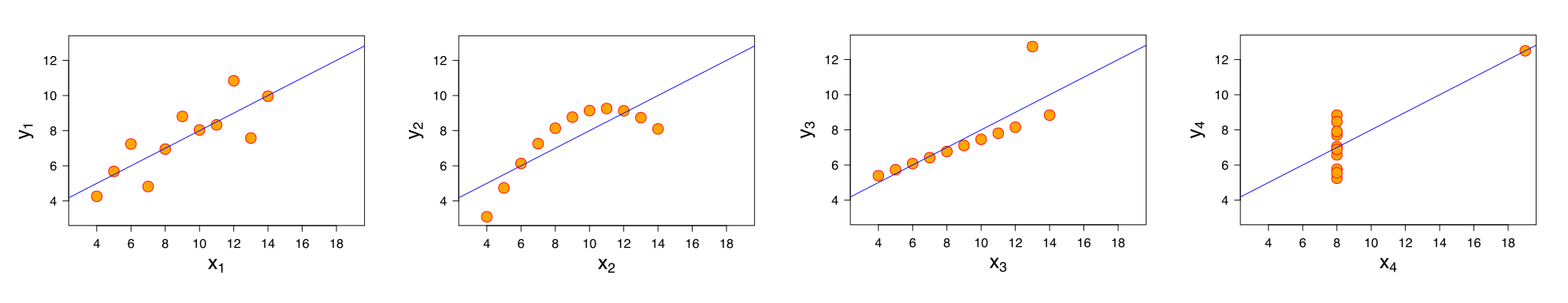}
    \caption{Anscombe's quartet \cite{a23325bf-6a64-3bba-a051-ebe1b2dce874}.}
    \label{fig:anscombe}
\end{figure}

\subsection{The Flaws of Interpretable Predictors}\label{sec:flaws_int}

\begin{intprob}\label{intprob_1}
    Interpretability can't provide all that there is to know about a predictor.
\end{intprob}

Anscombe's quartet \cite{a23325bf-6a64-3bba-a051-ebe1b2dce874} (see \autoref{fig:anscombe}) efficiently illustrates that many datasets, given a training algorithm, could lead to a same predictor: "[they] are designed to have approximately the same linear regression line (as well as nearly identical means, standard deviations, and correlations) but are graphically very different."\footnote{\url{https://en.wikipedia.org/wiki/Linear_regression}} This illustrates the pitfalls of relying solely on a fitted predictor for trying to understand the relationship between variables or the role of each of its parameters. 

Many other information aren't displayed by the predictor in itself: adversarial examples, or how is it possible to trick the decision process; influential instances, or how influential was a certain training example when training the predictor; feature interaction, or to what extent the prediction is the result of joint effects of the features; relative feature importance; counterfactual explanations, or how an instance has to change to significantly change its prediction; etc.

\newpage

\begin{figure}[htp]
    \centering
    \begin{subfigure}[b]{\textwidth}
    \centering
        \includegraphics[width=0.8\textwidth]{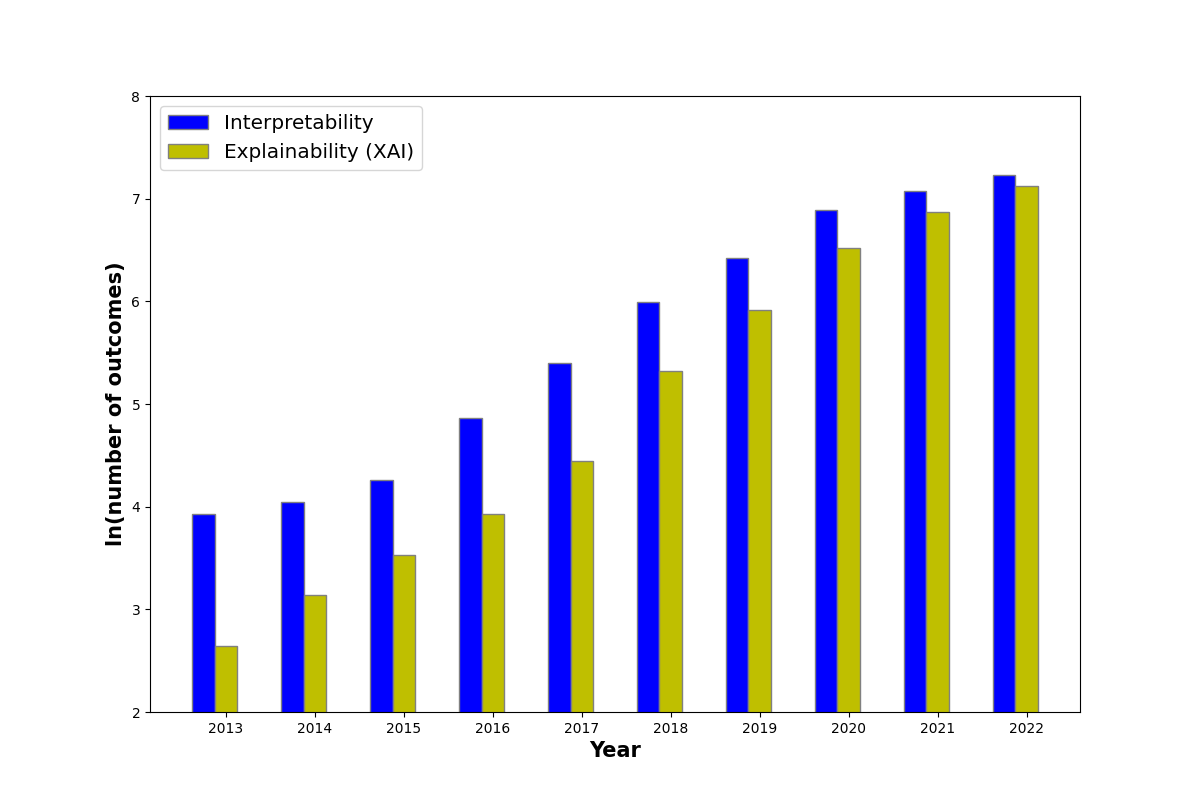}
    \caption{Comparison between the natural log of the number of outcomes for specific query searches on ArXiv as a function of the year. Searches were done for the Computer Science (cs) subject, with cross-listed papers. The query for Interpretability: ("Interpretability" $\lor$ "Interpretable" in Title) $\lor$ ("Interpretability machine learning" $\lor$ "Interpretable machine learning" in Abstract). Query for Explainability: ("Explainability" $\lor$ "Explainable"  $\lor$ "XAI" in Title) $\lor$ ("Explainability machine learning" $\lor$ "Explainable machine learning" $\lor$ "XAI" in Abstract). One must not forget that in many cases, "interpretability" refers to our definition of "explainability", but the inverse is not true.}
    \label{fig:int-exp_1}
    \end{subfigure}
    
    \begin{subfigure}[b]{\textwidth}
    \vspace{5mm}
        \centering
        \includegraphics[width=0.85\linewidth]{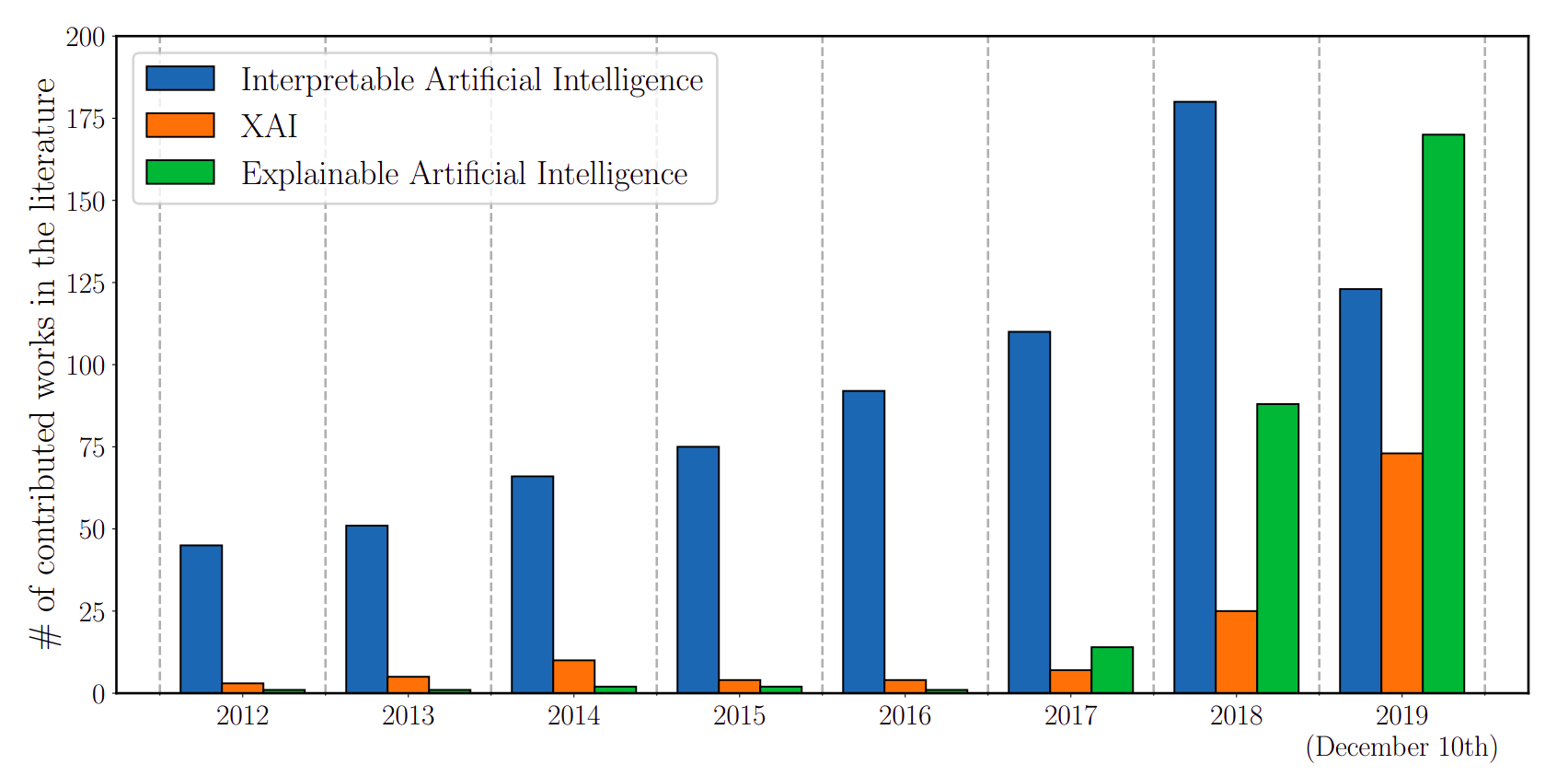}
        \caption{Borrowed from \citet{DBLP:journals/inffus/ArrietaRSBTBGGM20}: “Evolution of the number of total publications whose title, abstract and/or keywords refer to the field of XAI during the last years. Data retrieved from Scopus\textregistered~(December 10th, 2019) by using the search terms indicated in the legend when querying this database. [...]”}
        \label{fig:int-exp_2}
    \end{subfigure}
    
    \caption{Trends in interpretability and explainability (XAI) research popularity.}\label{fig:int-exp}
\end{figure}

\begin{intprob}\label{intprob_2}
    Interpretability is subjective.
\end{intprob}

Interpretability cannot be objectively computed, for it depends on both how the judge makes sense of the predictor and their preferences: it is an opinion. Therefore, two judges could have a completely different idea of the inner workings of a given predictor. And as for opinion, one can only argue in favor of his views on the interpretability of a predictor, but can't acquire a view on a predictor from someone else.

\begin{intprob}\label{intprob_3}
    Enforcing interpretability can make training harder.
\end{intprob}

Since the space of interpretable predictors is a subspace of the ensemble of every predictor, enforcing interpretability is a constraint to apply to an algorithm. In practice, the following are usually considered for favoring transparency: reducing the number of features of the predictor, the number of non-zero parameters of the predictor, or the number of bits required in order to encode some of the predictor's parameters. Such constraints can lead to delicate discrete optimization problems where making steps forward in efficiently handling those is not an easy task. For instance, when it comes to the efficient training of decision trees, CART \cite{DBLP:books/wa/BreimanFOS84} (which dates back to 1984) still is a state-of-the-art approach for training decision trees.

This might explain the trend presented on \autoref{fig:int-exp}, where it seems like there is a tendency, probably in agreement with the idea that explainability and interpretability are substitutes for one another, to neglect interpretable approaches and favoring the training of black-boxes before explaining them.

\begin{intprob}\label{intprob_4}
    Interpretability as a means is not enough to ensure some of its most common ends.
\end{intprob}

It is clear that interpretability is needed for many reasons: if algorithms lack transparency, domain experts or the public will not trust them \cite{Kim2015InteractiveAI}; people have a right to know why an ML algorithm has produced some verdict (such as lack of creditworthiness) about them \cite{Goodman_Flaxman_2017}; algorithms whose inner workings are not interpretable can't enable us to produce causal explanations of the world \cite{pittphilsci14452}; etc.

For example, fairness is an end to which interpretability is sometimes used as a mean \cite{doshivelez2017towards,molnar2022}, where the predictions need to be unbiased toward one or some sensitive variables. Not using such variables in the decision process simply isn't enough to ensure fairness because sometimes, information regarding the sensible variables is encapsulated in benign variables via complex relationships. Such relationships are unknown and might be numerous. For these reasons, the different aspects of fairness (e.g. independence, separation, sufficiency) are mostly only statistically demonstrated. Interpretability only permits the knowing of exactly how the benign variables are manipulated by a predictor, which simply is not enough to certify its fairness \cite{1548834}. This goes as well for privacy, where sensible variables would be personal information. 

Reliability, robustness or trust are also ends to which interpretability is sometimes used as a mean, each of them being seemingly connected: can we trust a predictor that is not reliable? Can it be reliable if it is not robust? The problem is that interpretability does not imply any of these characteristics. A predictor that is interpretable is not necessarily a predictor in which one can blindly put its trust \cite{DBLP:journals/corr/abs-2202-05302}: interpretability only yields an argument, which can be used to back the proposition that we should trust (or not) the predictor: it yields clarity, not a guarantee. That same predictor would not necessarily be robust or reliable either, but understanding the predictor and its training algorithm leads to insights into its limitations and weaknesses. In the end, interpretability induces confidence in how much a predictor \textit{can} be reliable, robust, or trusted.

\subsection{Explained Interpretable Predictor}\label{sec:reconc_int_exp}

Here, we make the case that interpretability and explainability are not substitutes for one another, but in many occasions complementary, in the sense that they reinforce one another. We do so by looking at all of the problems that have been raised in Subsection \ref{sec:flaws_exp} and Subsection \ref{sec:flaws_int}, but now when explainability and interpretability meet.

First, the blindness with which the explainability approaches must be trusted (\autoref{expprob_1}) substantially decreases, for some explanations can be more easily verified or corroborated by a look at the interpretable predictor itself. Interpretability permits us to justify whether or not to put our trust in an explanation. The same goes for the over-reliance on explainability. The over-reliance might exist partially because the explanations are the only resources we have for understanding what is going on within the predictor. With interpretable predictors, it is now easier either to be skeptical toward explanations or to justify this reliance.

The fact that each explainability technique is, to a certain extent, misaligned either with the explainer’s intent or the explainee’s (\autoref{expprob_3}) is also attenuated. With a black-box, the explanation has to be received and analyzed by the explainee with no reference whatsoever, or other explanations. With an interpretable predictor, there is an objective common ground on which both the explainer and the explainee can understand each other; indeed, having a reference point when it comes to transferring knowledge (explanation) makes the chances of a discrepancy between what was intended to share and what has been perceived less likely to happen, and of a smaller scale when it does happen \cite{DBLP:journals/ai/Miller19}. As for the discrepancies between explainability techniques, \citet{markus2023understanding} shows that for a decreasing number of features, there is a decreasing disagreement between different feature importance techniques. The number of considered features (parsimony) is only a facet of interpretability, but this might be a first step toward demonstrating that the more interpretable the predictor is, the fewer discrepancies there are between the explainer's intent and the explainee's interpretation of the information.

Even though explanations are always wrong (\autoref{expprob_2}), they are less wrong when it comes to interpretable predictors. Indeed, since explanations are simplifications of the original predictor, the simpler the original predictor (which in many cases is proportionate to the degree of interpretability), the lesser of an approximation is required by the explanation. Thus, the explanations might just be more faithful to the original predictor. In the same line of thought, with an interpretable predictor, there are many cases where the computational costs will be reduced, thus permitting the use of more computationally heavy but also more authentic methods \cite{DBLP:conf/nips/LundbergL17}.

In some cases, the explanations do raise questions themselves (\autoref{expprob_4}), but the interpretable predictor can answer those ones. For example, an attention-based technique \cite{DBLP:conf/cvpr/WangJQYLZWT17,DBLP:conf/recsys/SeoHYL17,DBLP:journals/access/ZhangW23} aims at identifying what input has been mostly important for rendering a given prediction, but raises the following question: “How did those inputs were actually used to render that prediction?” With an interpretable predictor, that later question inherently can be answered. We mentioned earlier in \autoref{expprob_4} the example from \citet{DBLP:journals/natmi/Rudin19} of real estate agents who are pricing a house, where “a \textit{black-box} real estate agent would provide the price with no explanation”, whereas “a saliency real estate agent would say that the price is determined from the roof and backyard, but does not explain how they were used to determine the price”. Well, in contrast, “an interpretable agent would explain the calculation in detail, for instance, using \textit{comps} or comparable properties to explain how the roof and backyard are comparable between properties, and how these comparisons were used to determine the price”.

Finally, simpler (parsimonious, sparse) predictors tend to be more robust toward adversarial attacks \cite{DBLP:conf/ijcnn/TabacofV16,DBLP:conf/eurosp/PapernotMJFCS16,DBLP:conf/sp/PapernotM0JS16,DBLP:journals/midm/RodriguezNCKH22}. For example, \citet{DBLP:conf/sp/PapernotM0JS16} propose a framework for distilling networks in order to make them more robust; one could think of a scenario where the distillation is made after a disentanglement layer in a deep neural network. \citet{DBLP:journals/midm/RodriguezNCKH22} shows similar conclusions, but this time in a practical context: when the inputs are medical images.

We made the case that interpretability can help reduce the flaws in explaining a predictor. Here, we make the case that the flaws of interpretability, when applied to explained predictors, are attenuated. 

Interpretability is not sufficient to provide information about the parameters and features of a predictor (\autoref{intprob_1}), whereas various explainability techniques can compute feature importance, distribution patterns, visual representation of the inputs/features and the predictor, statistics on parameters and inputs, etc. By doing so, humans don't have to let their intuition guide them in these matters, so it leads both the judges toward a single view on the predictor (objectivity replaces subjectivity, thus diminishing \autoref{intprob_2}).

Even though the training of interpretable predictors can be harder than the training of any kind of predictor (\autoref{intprob_3}), explainability, in its exploratory purpose, can come in handy: variable selection \cite{DBLP:journals/electronicmarkets/ZachariasZCH22,DBLP:journals/tvcg/KrausePB14,DBLP:conf/visualization/ZhaoKMWZCE19}, parameter pruning \cite{YEOM2021107899,jimaging8020030}, guiding the modeling of the predictor \cite{DBLP:journals/tnn/YangZS21} or understanding the causal relationship between variables \cite{pittphilsci18005} all can help during the training phase. For example, in order to favor parsimony, one could build its predictor in a top-down fashion but make use of SHAP values for deciding which features to remove.

Interpretability alone isn't enough to ensure some of its most common ends (\autoref{intprob_4}). When it comes to group fairness (independence, separation, sufficiency) or individual fairness, interpretability might ensure that the benign variables are used properly, but explainability only is able to compute metrics assessing true fairness \cite{DBLP:conf/aaai/Zhao0D23,DBLP:conf/ssci/StevensDVV20,DBLP:conf/icml/ZhouCH20} and provide many relevant statistics (feature importance, distribution patterns, etc.) for guaranteeing the reliability or trustworthiness of a predictor. Explainability can even be used in an adversarial fashion to find examples assessing the biases of a predictor.

\section{Conclusion}

We studied the major limitations of a predictor involving only interpretability or explainability and we conclude that considering both these aspects simultaneously leads to the mitigation of these many flaws: being less subjective in the interpretation of the predictor; ensuring that the predictor meets his ends; more reliable and truthful information; etc. Our analysis leads to a better understanding of these two abstract concepts in themselves and in their relationship. We invite the ML community to look at interpretability and explainability in a new way, and we hope it fosters ideas where both interpretability and explainability are met at the same time.

Similar to what motivated the present work, it would be important to put forth the inquiry on interpretability and explainability and empirically verify the benefits of being both interpretable and explainable on the following topics: robustness \cite{DBLP:conf/daset/HiaKP22,DBLP:journals/phycomm/SaengsawangL23,DBLP:journals/corr/abs-2303-06199}, algorithmic stability \cite{DBLP:conf/alt/RajBGZS23,DBLP:conf/aaai/TamarSZ22,DBLP:journals/corr/abs-2111-15546}, fairness \cite{DBLP:journals/aiethics/AgarwalAA23,DBLP:journals/jair/FabrisEMS23,DBLP:journals/kais/ZhangZHCLZY23} / unbiasedness \cite{DBLP:journals/tcs/FriedrichKK19,DBLP:conf/pcs/AmraniSM16,DBLP:journals/ijsysc/ZhaoHML22}, private life/data privacy \cite{DBLP:books/sp/YuC23,DBLP:journals/access/ZhangZGS23,DBLP:journals/cn/ZhouZWHHGJ23}, etc.

% \bibliography{references}
% \bibliographystyle{icml2023}
\bibliographystyle{splncs04}
\bibliography{references}

% \end{document}

%%%%%%%%%%%%%%%%%%%%%%%%%%%%%%%%%%%%%%%%%%%%%%%%%%%%%%%%%%%%%%%%%%%%%%%%%%%%%%%
%%%%%%%%%%%%%%%%%%%%%%%%%%%%%%%%%%%%%%%%%%%%%%%%%%%%%%%%%%%%%%%%%%%%%%%%%%%%%%%
% APPENDIX
%%%%%%%%%%%%%%%%%%%%%%%%%%%%%%%%%%%%%%%%%%%%%%%%%%%%%%%%%%%%%%%%%%%%%%%%%%%%%%%
%%%%%%%%%%%%%%%%%%%%%%%%%%%%%%%%%%%%%%%%%%%%%%%%%%%%%%%%%%%%%%%%%%%%%%%%%%%%%%%
%\newpage
%\appendix
% \onecolumn

%\addcontentsline{toc}{section}{Appendix} % Add the appendix text to the document TOC
%\addcontentsline{toc}{section}{Appendix} % Add the appendix text to the document TOC
%\part{Appendix} % Start the appendix part
%\parttoc % Insert the appendix TOC
%\parttoc
% Ca serait pas mal d'insérer une table des matières pour aider le lecteur à naviguer dans les annexes mais je n'y parvies pas ;)

%%%%%%%%%%%%%%%%%%%%%%%%%%%%%%%%%%%%%%%%%%%%%%%%%%%%%%%%%%%%%%%%%%%%%%%%%%%%%%%
%%%%%%%%%%%%%%%%%%%%%%%%%%%%%%%%%%%%%%%%%%%%%%%%%%%%%%%%%%%%%%%%%%%%%%%%%%%%%%%

\end{document}